# CLIP-Joint-Detect: End-to-End Joint Training of Object Detectors with Contrastive Vision-Language Supervision

Behnam Raoufi*
*Electrical Engineering Department*
*Sharif University of Technology*
Tehran, Iran
behnam.raoufi93@sharif.edu

Hossein Sharify
*Electrical Engineering Department*
*Sharif University of Technology*
Tehran, Iran
sharifir.hossein@ee.sharif.edu

Mohamad Mahdee Ramezanee
*Electrical Engineering Department*
*Sharif University of Technology*
Tehran, Iran
ramezani.mm@ee.sharif.edu

Khosrow Hajsadeghi
*Electrical Engineering Department*
*Sharif University of Technology*
Tehran, Iran
ksadeghi@sharif.edu

Saeed Bagheri Shouraki
*Electrical Engineering Department*
*Sharif University of Technology*
Tehran, Iran
bagheri-s@sharif.edu

***Abstract*—Conventional object detectors rely on cross-entropy classification, which can be vulnerable to class imbalance and label noise. We propose CLIP-Joint-Detect, a simple and detector-agnostic framework that integrates CLIP-style contrastive vision-language supervision through end-to-end joint training. A lightweight parallel head projects region or grid features into the CLIP embedding space and aligns them with learnable class-specific text embeddings via InfoNCE contrastive loss and an auxiliary loss term, while all standard detection losses are optimized simultaneously. The approach applies seamlessly to both two-stage and one-stage architectures. We validate it on Pascal VOC 2007+2012 using Faster R-CNN and on the large-scale MS COCO 2017 benchmark using modern YOLO detectors (YOLOv11), achieving consistent and substantial improvements while preserving real-time inference speed. Extensive experiments and ablations demonstrate that joint optimization with learnable text embeddings markedly enhances closed-set detection performance across diverse architectures and datasets.**

## I. Introduction

Object detection is a core task in computer vision, enabling critical applications in autonomous driving, video surveillance, robotics, and intelligent image understanding. Two-stage detectors such as Faster R-CNN [1] have long dominated closed-set benchmarks like Pascal VOC [2] and MS COCO [3], delivering excellent localization and classification accuracy through powerful feature pyramid networks and region-based prediction heads. Nevertheless, these models rely exclusively on categorical cross-entropy classification, which can be sensitive to label noise, class imbalance, and subtle intra-class variations, often limiting further gains on challenging datasets. The introduction of Contrastive Language-Image Pre-training (CLIP) [4] has provided a new source of rich semantic supervision by learning highly robust visual representations aligned with textual descriptions on web-scale data.

In this paper, we propose CLIP-Joint-Detect, a simple yet effective framework that truly unifies mainstream object detectors with CLIP-style semantic supervision through end-to-end joint training. As illustrated in Fig. 1, we attach a lightweight parallel branch to the region (or grid) features of any detector [5]: a small projection head maps visual features into the CLIP embedding space, where they are aligned with learnable class-specific text embeddings via an InfoNCE contrastive loss and an auxiliary loss term.

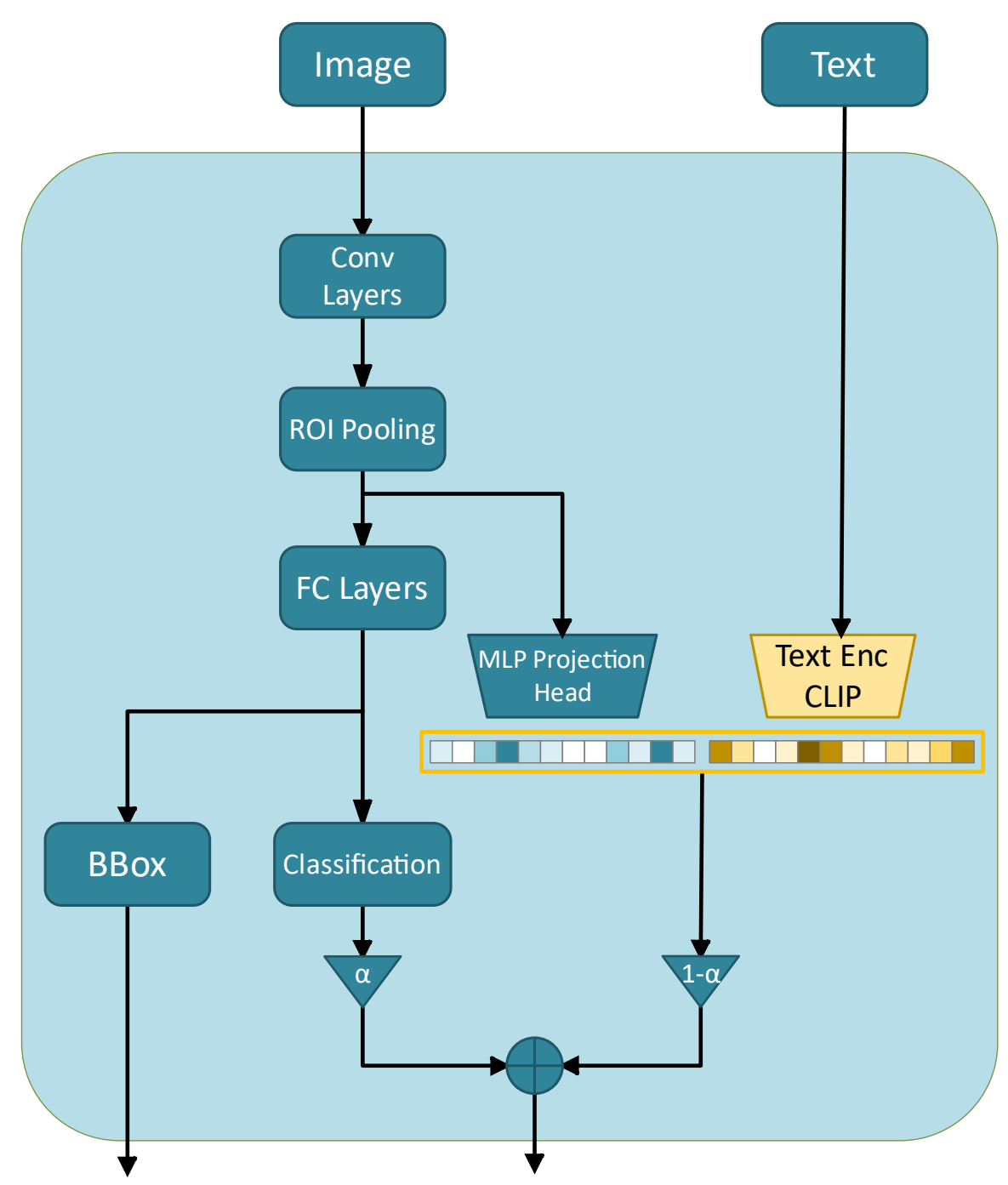

Fig. 1. The overall architecture of CLIP-Joint-Detect.

All objectives—including standard bounding-box regression, objectness, and classification losses—are optimized simultaneously, allowing gradients from the vision-language alignment task to directly improve the shared backbone and proposal generation mechanism.

## II. Related work

The modern era of object detection began with the R-CNN [6] architecture, which established the two-stage paradigm by extracting region proposals followed by powerful region-wise classification and dramatically raised accuracy on Pascal VOC. Fast R-CNN introduced RoI pooling to eliminate redundant computations, while Faster R-CNN integrated a fully learnable Region Proposal Network (RPN), enabling true end-to-end training and becoming the backbone of virtually all subsequent high-accuracy detectors. This line of work has since been extended in numerous directions: R-FCN removed heavy per-region sub-networks, Mask R-CNN [7] added instance segmentation, Feature Pyramid Networks (FPN) [8] and its successors (PANet [9], NAS-FPN [10], BiFPN [11]) addressed scale variation through multi-level feature fusion, Cascade R-CNN [12] and HTC introduced multi-stage refinement for progressively better bounding boxes and features, and DyHead [13] unified classification, regression and feature aggregation in a dynamic head.

Parallel to these developments, one-stage detectors emerged to achieve much higher efficiency. YOLO [14] reframed detection as a single-shot regression problem and enabled real-time inference, SSD adopted multi-scale feature maps for dense predictions, RetinaNet introduced focal loss to overcome extreme foreground-background imbalance, and the transformer-based DETR family eliminated hand-crafted components entirely in favour of attention-based set prediction. The YOLO series [15] has remained the most widely deployed real-time detector family, evolving steadily through YOLOv5, YOLOv8, YOLOv10, and YOLOv11 [16], consistently pushing the speed-accuracy frontier for practical applications.

All dominant detection architectures, whether two-stage or one-stage, rely almost exclusively on categorical cross-entropy (often with focal loss or label-smoothing variants) for classification supervision. This has proven highly effective on standard closed-vocabulary benchmarks but is inherently limited when confronted with severe class imbalance, noisy annotations, or large intra-class variation, particularly for small, occluded, or long-tail objects.

The introduction of large-scale contrastive vision-language pre-training with CLIP and its many improved variants (OpenCLIP, SigLIP, EVA-CLIP, InternViT-CLIP, etc.) provided a qualitatively different and far more semantic form of supervision. While the vast majority of subsequent detection research has exploited CLIP-style models to enable open-vocabulary or zero-shot detection (ViLD [17], RegionCLIP [18], OWL-ViT, GLIP/GLIPv2 [19], Grounding DINO [20], YOLO-World [21], etc.), a few works have explored their use for conventional closed-vocabulary tasks, typically via frozen-encoder distillation or feature enrichment, resulting in only small improvements of ∼1–2 AP on MS COCO.

Direct end-to-end joint training of standard object detectors together with CLIP-style contrastive vision-language objectives and learnable class-specific text embeddings remains largely unexplored in the closed-vocabulary setting.

## III. PROPOSED METHOD

### A. Overview

We propose CLIP-Joint-Detect, a unified training framework that incorporates CLIP-style contrastive vision-language supervision directly into any standard object detector through genuine end-to-end joint optimization. Unlike prior decoupled or multi-stage methods, our approach attaches a lightweight parallel head to the detector's region or grid features and trains all components—localization, conventional classification, and vision-language alignment—simultaneously using a single composite loss. The method is fully architecture-agnostic and has been successfully applied to both two-stage (Faster R-CNN) and one-stage (YOLOv11) detectors without modification of their original training pipelines.

### B. Base Detector

Our method uses the feature vector that every modern detector already creates for each possible object right before its final prediction layers.

For two-stage detectors such as Faster R-CNN with ResNet-50-FPN, the backbone extracts multi-scale feature maps, the Region Proposal Network (RPN) generates candidate proposals, and RoI-Align produces fixed-size 7×7×256 region features. These are globally average-pooled (and optionally passed through a short linear head) to yield a compact feature vector $f_i \in \mathbb{R}^{1024}$ that feeds the standard box regressor and (C+1)-way classifier.

For one-stage detectors such as YOLOv11, the backbone and neck generate multi-scale feature maps, and the detection head directly outputs a feature vector $f_i \in \mathbb{R}^{256}$ or $\mathbb{R}^{512}$ for each anchor or grid cell that is used by independent linear layers to predict bounding-box offsets, objectness score, and C-way class probabilities.

In both cases, $f_i$ is the exact high-level representation already consumed by the original detector. Our CLIP-guided branch is attached to this identical shared $f_i$.

### C. CLIP-Guided Parallel Head

From each shared feature vector we add a small parallel branch consisting of a two-layer MLP that projects the feature into a 512-dimensional visual embedding, followed by L2 normalization (so every visual embedding has unit length).

We maintain one trainable 512-dimensional text embedding for each dataset class (20 classes for Pascal VOC, 80 for COCO). These text embeddings are initialized by feeding simple prompts such as "a photo of a {class_name}" into the public CLIP text encoder (ViT-B/32), but they are fully updated during training.

The similarity between a normalized visual embedding $\hat{v}_i$ and a text embedding $t_c$ is calculated with a scaled dot product:

$$s(\hat{v}_i, t_c) = \frac{\hat{v}_i{}^T t_c}{\tau_c} \quad (1)$$

where $\tau_c$ is a learnable class-specific temperature parameter — a separate scalar for each dataset class ($\tau_c \in R^c$, C = 20 for Pascal VOC, C = 80 for MS COCO), independently initialized to 0.07.

This design means that every class has its own dedicated temperature value that directly controls the scaling of only its own similarity scores across all visual embeddings during both training and inference. While the visual-text dot product $\hat{v}_i{}^T\ tc$ remains identical for a given region and class regardless of the temperature formulation, dividing by a per-class $\tau_c$ allows each category to independently modulate how sharply or softly its similarity contributes to the final softmax probability distribution in both the InfoNCE contrastive loss and the auxiliary loss term.

During training, all $\tau_c$ values are fully differentiable parameters updated end-to-end via standard back-propagation alongside the projection MLP, the text embeddings $t_c$, and the rest of the detector. At inference, the learned per-class $\tau_c$ values are used exactly as during training to compute the CLIP-branch probabilities via $p_c{}^{clip} = softmax(s_i)$ ,

ensuring perfect consistency between training and evaluation objectives.

Only positive samples (IoU ≥ 0.5 with any ground-truth box) contribute to the vision-language losses. The total training loss is:

$$L_{total} = L_{det} + \lambda_{cont} L_{cont} + \lambda_{aux} L_{aux} \quad (2)$$

where $L_{det}$ is the original full loss of the detector, and $\lambda_{cont}$, $\lambda_{aux}$ are small balancing weights. The main vision-language objective is the symmetric InfoNCE contrastive loss:

$$L_{cont} = -\frac{1}{2}\left(L_{i2t} + L_{t2i}\right) \quad (3)$$

with the image-to-text component:

$$L_{i2t} = -\frac{1}{N}\sum_{i}\log\frac{\exp\left(s(v_i, t_{y_i})\right)}{\sum_c \exp(s(v_i, t_c))} \quad (4)$$

and a symmetric text-to-image term $L_{t2i}$. Here $y_i$ is the ground-truth class of sample $s_i$, and $\tau$ is a learnable temperature parameter initialized to 0.07. An auxiliary sigmoid loss is added for stability and faster convergence:

$$L_{aux} = \frac{1}{N}\sum_{i}\left[-\log\left(\sigma\left(t.s\left(v_i, t_{y_i}\right) + b\right)\right) + \frac{1}{M_i}\sum_{j \neq y_i}^{M_i}\log\left(\sigma\left(-t.s\left(v_i, t_j\right) - b\right)\right)\right] \quad (5)$$

where $\sigma$ is the sigmoid, $t > 0$ is learnable inverse temperature (init. 1.0), b is learnable bias (init. 0), and $M_i$ is the number of negatives per positive. all parameters — backbone, original detection heads, the projection MLP, the class text embeddings $w_c$ and the temperature $\tau$ — are updated jointly in every training step.

## D. *Inference and Score Fusion*

At inference, the base detector outputs its standard class probabilities ${p_c}^{ce}$. The CLIP branch produces similarity-based probabilities ${p_c}^{clip} = softmax(s_i)$. The final confidence for each class is obtained by weighted fusion:

$$score_c = \alpha . {p_c}^{ce} + (1-\alpha) . {p_c}^{clip} \quad (6)$$

where the value 0.7 was chosen for $\alpha$ after evaluation on the validation set and is used consistently for all reported results on both Pascal VOC and MS COCO. Non-maximum suppression is subsequently applied using the fused scores [22]. The auxiliary CLIP branch incurs negligible computational cost and may be discarded entirely at test time, preserving the original detector's inference speed [23].

# IV. EXPERIMENTS

## A. *Datasets and Evaluation Protocols*

We evaluate CLIP-Joint-Detect on two standard object detection benchmarks.

1) Pascal VOC 2007+2012:

Following standard practice, we train on the combined VOC 2007 trainval and 2012 trainval sets (16,551 images, 20 classes) and evaluate on the VOC 2012 test set (10,991 images). Our method is integrated into Faster R-CNN, and performance is reported as mean Average Precision at IoU=0.5 (mAP@0.5).

2) MS COCO 2017:

We train on the train2017 split (118,287 images, 80 categories) and evaluate on the val2017 split (5,000 images). Our approach is incorporated into YOLOv11, and performance is reported using the primary COCO metric: mAP@[0.5:0.95].

## B. *Implementation Details*

We implement CLIP-Joint-Detect atop the official torchvision Faster R-CNN ResNet-50-FPN and Ultralytics YOLOv11n/YOLOv11s codebases. The CLIP projection head is a two-layer MLP. Class text embeddings are initialized via OpenAI CLIP ViT-B/32 with the prompt "a photo of a {class_name}" and remain learnable, along with a scalar temperature $\tau$ (init. 0.07). Faster R-CNN uses a frozen backbone, SGD (lr=0.005, momentum=0.9, wd=1e-4), batch size 16, and StepLR ($\gamma = 0.1$ every 3 epochs); YOLOv11 follow their default AdamW optimizer and schedules with batch size 64. Vision-language losses are weighted by $\lambda_cont = 0.5$ and $\lambda_aux$=0.8. Inference applies score fusion with $\alpha = 0.7$. No test-time augmentation or ensemble is used.

## C. *Main Results*

1) Pascal VOC 2012

As shown in Fig. 2, CLIP-Joint-Detect significantly outperforms the vanilla Faster R-CNN baseline, where per-class AP@0.5 bars on Pascal VOC 2012 val reveal consistent improvements—especially large on difficult classes like bottle, potted plant, chair, and person—with the extension beyond the baseline segment representing our method's contribution. These qualitative gains are corroborated by Fig. 3, which reliably demonstrates detection and classification of challenging small or heavily occluded objects that the baseline misses or mislabels, while maintaining localization precision. These qualitative observations are fully consistent with the large quantitative gains reported in Table I.

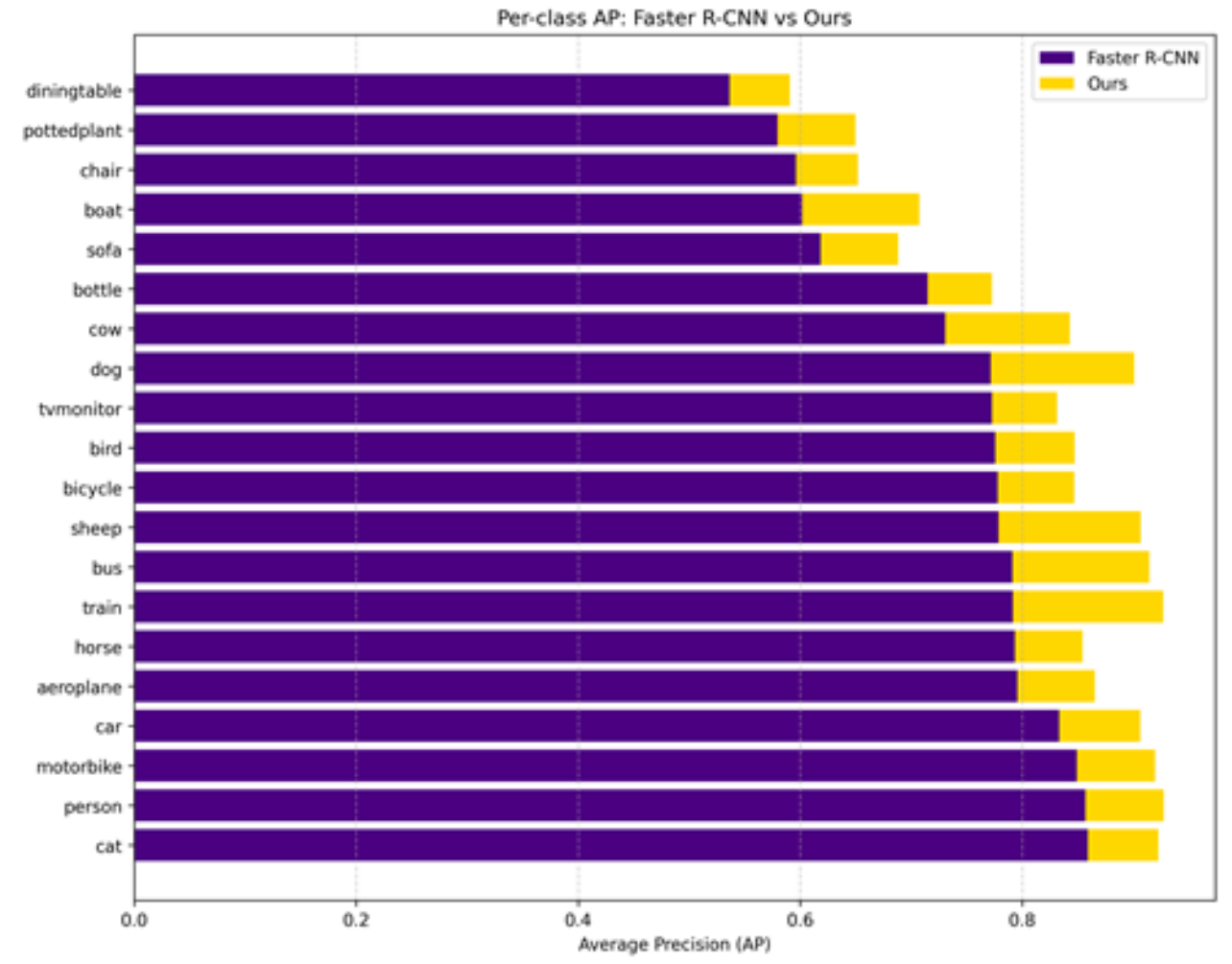

Fig. 2. Per-class AP@0.5 comparison on Pascal VOC 2012 val set. Each bar represents the performance of CLIP-Joint-Detect, with the initial segment showing the AP of the vanilla Faster R-CNN ResNet-50-FPN baseline and the extension indicating the additional gain from our method. The total bar length is therefore the final AP achieved by CLIP-Joint-Detect.

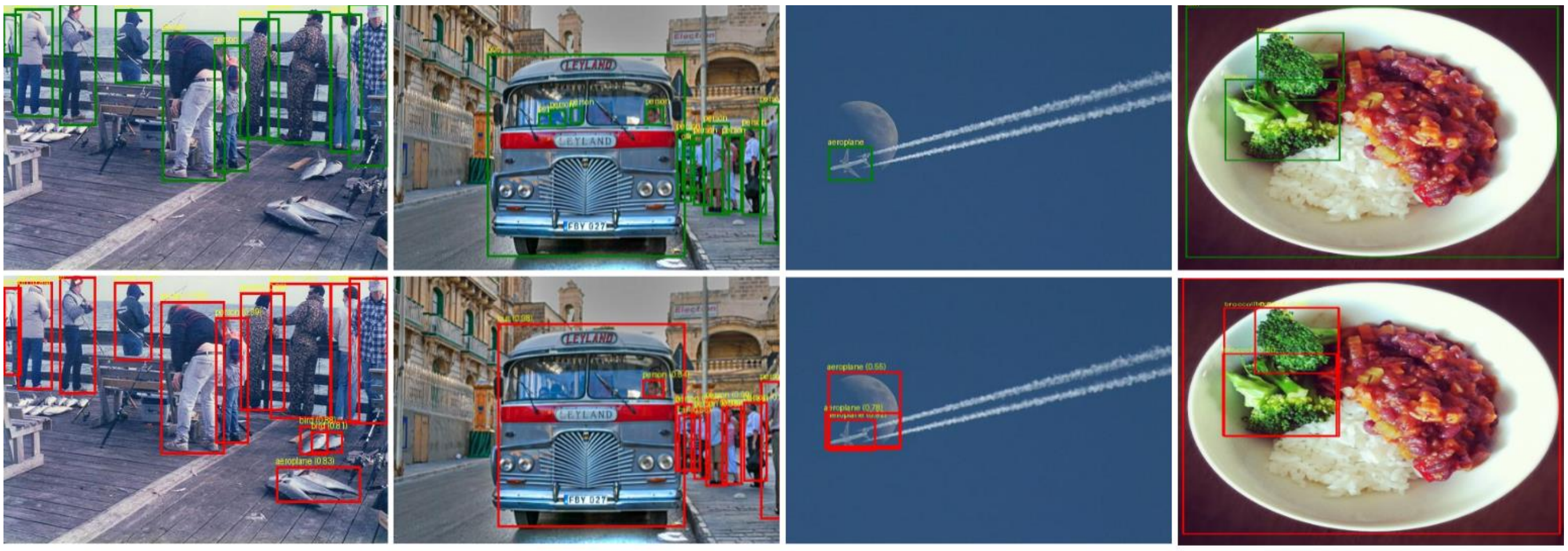

Fig. 3. Qualitative comparison on Pascal VOC 2012 val images (top: CLIP-Joint-Detect; bottom: vanilla Faster R-CNN, same confidence threshold)

Our method achieves an absolute improvement of +5.3 points over the strong torchvision baseline and surpasses recent specialized detection heads while retaining the original simple RoI architecture.

2) MS COCO 2017:

As clearly shown in Fig. 4, the vanilla YOLOv11S frequently misses or misclassifies small objects and instances that are partially overlapped or occluded by larger objects. These cases are particularly challenging for conventional detectors because their purely visual feature representations tend to be dominated by salient foreground regions or by the texture of larger nearby objects. In contrast, our CLIP-Joint-Detect method successfully detects and correctly labels these difficult cases, significantly reducing both false negatives and classification errors while maintaining precise bounding boxes.This improvement directly stems from the richer semantic representations learned through joint vision-language training: the detector is encouraged not only to localize objects but also to align their visual features with class-specific textual embeddings that encode higher-level conceptual cues such as shape, context, and co-occurrence patterns. As a result, ambiguous or low-resolution regions receive stronger class-discriminative signals, enabling more robust predictions under occlusion, clutter, and scale variation. We apply the exact same CLIP-Joint-Detect training recipe to the entire YOLOv11 family (Nano, Small, Medium, and Large variants) on MS COCO 2017 without any architecture-specific tuning. As summarized in Table II, our method yields consistent and significant gains across all model sizes, demonstrating both scalability and architectural robustness.

TABLE I. compares CLIP-Joint-Detect (based on Faster R-CNN with ResNet-50) against the vanilla baseline and representative prior methods on the Pascal VOC 2012 test set using the standard mAP@0.5 metric.

| Model | Backbone | Map@0.5 |
|---|---|---|
| Faster R-CNN | VGG-16 | 72.3 |
| Faster R-CNN | ResNet-101 | 76.4 |
| Faster R-CNN | ResNet-50 | 74.13 |
| Ours | ResNet-50 | **81.7** |

For lightweight variants such as YOLOv11-Nano, the incorporation of contrastive vision-language supervision compensates for the limited feature capacity and leads to disproportionately large improvements on small and medium objects. For the larger YOLOv11-Medium and YOLOv11-Large models, the gains manifest primarily in fine-grained class discrimination and reduction of long-tail classification errors. Importantly, the added parallel CLIP-guided head introduces only a negligible number of parameters (<0.5% overhead) and imposes no measurable latency during inference, since it is used only during training. This preserves real-time detection speed while delivering substantial improvements in accuracy, confirming that CLIP-based semantic alignment can be integrated into modern detectors with minimal computational cost.

The gains are now even more pronounced (+2.9 to +3.7 points) across the entire model scale spectrum, with the lightweight Nano and Small variants benefiting the most in relative terms. This makes CLIP-Joint-Detect especially valuable for resource-constrained and real-time applications. The results on two fundamentally different detector families and dataset scales strongly confirm the generality and scalability of our joint vision-language training paradigm.

TABLE II. Performance on MS COCO 2017 val

| Method | Model Size | mAP@[0.5:0.95] | Gain |
|---|---|---|---|
| YOLOv11-N (baseline) | Nano | 39.5 | - |
| CLIP-Joint-Detect (YOLOv11-N) | Nano | 43.2 | **+3.7** |
| YOLOv11-S (baseline) | Small | 47.0 | - |
| CLIP-Joint-Detect (YOLOv11-S) | Small | 50.6 | **+3.6** |
| YOLOv11-M (baseline) | Medium | 51.5 | - |
| CLIP-Joint-Detect (YOLOv11-M) | Medium | 55.1 | **+3.6** |
| YOLOv11-L (baseline) | Large | 53.4 | - |
| CLIP-Joint-Detect (YOLOv11-L) | Large | 56.4 | **+2.9** |

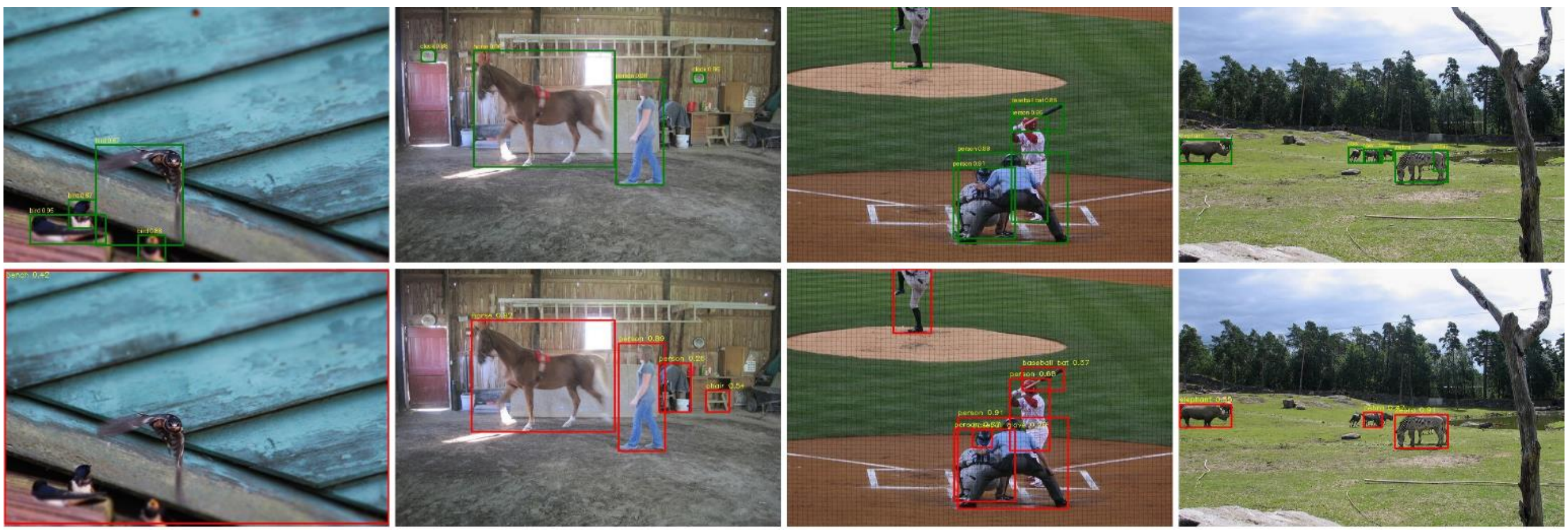

Fig. 4 Qualitative comparison on challenging MS COCO 2017 val images (top: CLIP-Joint-Detect with YOLOv11-S; bottom: vanilla YOLOv11-S, same confidence threshold).

## V. CONCLUSIONS

In this paper, we presented CLIP-Joint-Detect, a simple yet highly effective framework that integrates CLIP-style contrastive vision-language supervision directly into arbitrary object detectors through genuine end-to-end joint training. By attaching a lightweight parallel branch to the detector's existing region or grid features and optimizing a composite loss comprising standard detection objectives, an InfoNCE contrastive loss, and an auxiliary loss—all with fully learnable class text embeddings—our approach enables the backbone and proposal mechanism to produce representations that are inherently better aligned with rich semantic knowledge from large-scale pre-trained vision-language models. At inference, a simple late fusion of the original detector scores and CLIP similarity scores further boosts performance.

Extensive experiments on Pascal VOC 2012 and MS COCO 2017 demonstrate the effectiveness and generality of CLIP-Joint-Detect. On Pascal VOC, our method turns a standard Faster R-CNN ResNet-50-FPN into a detector that significantly outperforms both the vanilla baseline and recent specialized detection heads. On the much larger and more diverse MS COCO benchmark, applying the exact same recipe to the entire YOLOv11 family (Nano through Large) yields consistent absolute gains of 2.9 to 3.7 points across all model sizes while introducing negligible computational overhead and preserving real-time speed. Qualitative results further confirm that the gains are particularly pronounced on small, occluded, or semantically challenging objects—cases where traditional cross-entropy supervision alone often struggles. These results establish CLIP-Joint-Detect as a versatile, detector-agnostic paradigm that brings substantial closed-set performance improvements with minimal architectural changes and no additional test-time complexity.

## VI. FUTURE WORK

Although CLIP-Joint-Detect already achieves strong closed-set performance with minimal overhead, several natural extensions remain underexplored: scaling the framework to larger vision-language backbones (e.g., ViT-L/14 or modern open-source equivalents) to further boost accuracy on fine-grained categories; applying the same joint-training recipe to dense-prediction tasks such as instance segmentation and keypoint estimation, where semantic alignment could provide similar benefits; extending the approach to 3D object detection on point-cloud or multi-view inputs; and developing lightweight knowledge-distillation techniques to transfer the learned CLIP-style knowledge into the main detector head, allowing complete removal of the auxiliary branch after training for even faster inference on resource-constrained devices. These directions leverage the simplicity and generality of our paradigm while promising broader impact across computer-vision tasks.